\documentclass[sigconf]{acmart}

\acmConference[SIGKDD '25]{The 31-th ACM SIGKDD Conference on Knowledge Discovery and Data Mining (KDD '25)}{Oct 19 – 23th, 2025}{Honolulu, Hawai'i}

\acmSubmissionID{2275}

\usepackage{amsmath} 
\usepackage{booktabs}
\usepackage{multirow}
\usepackage[shortlabels]{enumitem}

\copyrightyear{2025}
\acmYear{2025}
\setcopyright{acmlicensed}\acmConference[KDD '25]{Proceedings of the 31st ACM SIGKDD Conference on Knowledge Discovery and Data Mining V.2}{August 3--7, 2025}{Toronto, ON, Canada}
\acmBooktitle{Proceedings of the 31st ACM SIGKDD Conference on Knowledge Discovery and Data Mining V.2 (KDD '25), August 3--7, 2025, Toronto, ON, Canada}
\acmDOI{10.1145/3711896.3736971}
\acmISBN{979-8-4007-1454-2/2025/08}

\begin{document}

\title[Precipitation Retrieval with Multimodal Knowledge Expansion]{From Swath to Full-Disc: Advancing Precipitation Retrieval\\ with Multimodal Knowledge Expansion}

\author{Zheng Wang}
\email{zhengwang@zjut.edu.cn}
\author{Kai Ying}
\email{221123120165@zjut.edu.cn}
\affiliation{%
  \institution{College of Computer Science, Zhejiang University of Technology}
  \city{Hangzhou}
  \state{Zhejiang}
  \country{China}
}

 \author{Bin Xu}
 \email{xubin@cma.gov.cn}
 \author{Chunjiao Wang}
 \email{wangcj@cma.gov.cn}
\affiliation{%
  \institution{National Meteorological Information Center}
  \city{Beijing}
  \country{China}
}

 \author{Cong Bai $^ \dag$}
\thanks{\dag Corresponding author}
\email{congbai@zjut.edu.cn}
\affiliation{%
  \institution{College of Computer Science, Zhejiang University of Technology\& Zhejiang Key Laboratory of Visual Information Intelligent Processing}
  \city{Hangzhou}
  \state{Zhejiang}
  \country{China}
}

\begin{abstract}
Accurate near-real-time precipitation retrieval has been enhanced by satellite-based technologies. However, infrared-based algorithms have low accuracy due to weak relations with surface precipitation, whereas passive microwave and radar-based methods are more accurate but limited in range.
This challenge motivates the Precipitation Retrieval Expansion (PRE) task, which aims to enable accurate, infrared-based full-disc precipitation retrievals beyond the scanning swath. 
We introduce Multimodal Knowledge Expansion, a two-stage pipeline with the proposed PRE-Net model. In the Swath-Distilling stage, PRE-Net transfers knowledge from a multimodal data integration model to an infrared-based model within the scanning swath via Coordinated Masking and Wavelet Enhancement (CoMWE). 
In the Full-Disc Adaptation stage, Self-MaskTune refines predictions across the full disc by balancing multimodal and full-disc infrared knowledge.
Experiments on the introduced PRE benchmark demonstrate that PRE-Net significantly advanced precipitation retrieval performance, outperforming leading products like PERSIANN-CCS, PDIR, and IMERG.
The code will be available at \url{https://github.com/Zjut-MultimediaPlus/PRE-Net}.
\end{abstract}





\maketitle
\begin{figure}
    \centering
    \includegraphics[width=1\linewidth]{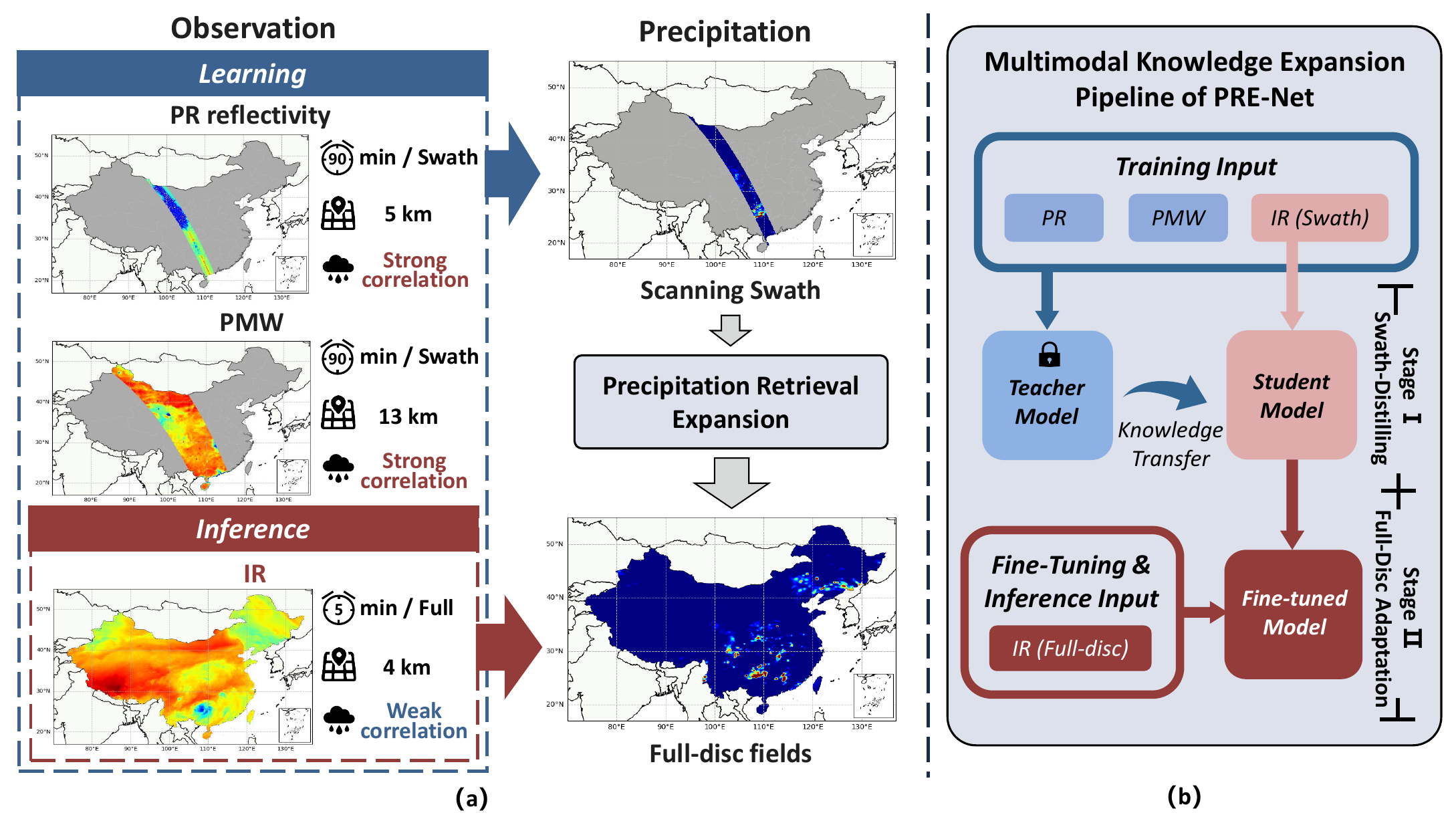}
    \caption{(a) The Precipitation Retrieval Expansion (PRE) task utilizes both strongly correlated PMW and PR data, as well as weakly correlated IR data for learning, while only IR data offers full-disc coverage. The PRE task leverages these heterogeneous data to achieve high-quality full-disc precipitation retrieval.
    (b) Multimodal knowledge expansion pipeline of PRE-Net. Multimodal knowledge is transferred via knowledge distillation in the Swath-Distilling stage and is further expanded from scanning swath to full-disc precipitation retrieval in the Full-Disc Adaptation stage.
    }
    \label{fig:fig1}
\end{figure}

\section{Introduction}

Accurate precipitation measurements are crucial for hydrometeorological studies, water resource management, and disaster prevention \cite{1,2}. These are typically divided into direct observations via ground-based rain gauges and indirect satellite retrievals. While rain gauges are reliable, their sparse distribution and inability to capture oceanic precipitation limit their effectiveness \cite{3}. In contrast, satellite-based retrievals offer broader coverage, making them a promising alternative for precipitation estimation.

Satellite-based precipitation retrieval primarily uses three types of sensor data: infrared (IR), passive microwave (PMW), and spaceborne precipitation radar (PR). 
IR data, collected by Geostationary-Earth-orbit (GEO) satellites, offers high spatiotemporal resolution ($\leq$4 km, 5-30 min intervals) and near-global coverage~\cite{lebedev2019precipitation}, making it widely used in operational retrievals~\cite{7,8,9}. 
However, IR data often shows a weak correlation with precipitation, as it mainly captures cloud-top features rather than hydrometeor profiles \cite{4,5}, leading to low-quality precipitation retrieval.
In contrast, PMW and PR reflectivity, obtained from low-Earth-orbit (LEO) satellites, can penetrate cloud layers to detect the 3D microstructure of precipitation particles~\cite{11}.
Advanced retrieval methods \cite{12,18,pmwdeeplearn2} exploit these direct precipitation-cloud relations, 
achieving superior performance. 
However, both PMW and PR suffer limited spatial coverage (narrow scanning swath of 2,873 km /245 km width), and low spatiotemporal resolution (90 mins, 13 km \& 90 mins, 5 km ) compared to IR data (full-disc, 5 mins, and 4 km). 
Those data's heterogeneous issue hinders their ability to offer continuous, global coverage, which is crucial for effective precipitation monitoring.

To overcome these limitations and enable high-quality full-disc precipitation retrievals, we propose a novel \textbf{P}recipitation \textbf{R}etrieval \textbf{E}xpansion \textbf{(PRE)} task, as illustrated in Fig.\ref{fig:fig1}(a). 
This task leverages the complementary strengths of multimodal data (IR, PMW, and PR) to address the constraints of narrow coverage and enhance retrieval accuracy. While multimodal data is available within a single scanning swath, IR remains the only modality with full-disc coverage, making it essential for expanding precipitation retrieval beyond the scanning swath.
This paper frames the PRE task as an \emph{incomplete multimodal learning} problem \cite{conf_kdd_WangZTZ20, conf_kdd_ChenZ20}.
To tackle this challenge, one possible approach is to generate missing multimodal data~\cite{hgan, cycae, smil}.  
However, the missing modalities are hardly recoverable due to the varying physical properties captured by the different data mentioned above.
Alternatively, modality-agnostic learning methods~\cite{mmformer, hemis, hetore} aim to learn representations that are invariant to the specific modality.
However, learning shared features may be useless when there is a strong imbalance in how each modality correlates with precipitation.
In contrast, we find that knowledge distillation (KD)\cite{mmanet, prototype, mlim} offers a more feasible solution. 
It transfers precipitation-related knowledge from high-quality multimodal data to low-quality IR data through a teacher-student model, facilitating more effective learning with missing modalities.
Despite this, transferring knowledge is challenging, especially because IR data’s broader coverage doesn’t necessarily translate to more useful knowledge about precipitation.

To answer the above challenges, we propose a \textbf{multimodal knowledge expansion} pipeline consisting of two stages, as illustrated in Fig.~\ref{fig:fig1}.
In the first stage, \textbf{Swath-Distilling}, we employ feature-based knowledge distillation to transfer precipitation-related knowledge from a teacher, trained on multimodal data (PMW, PR, IR), to a student that operates on IR-only.
We design a lightweight PRE-Net model with only 48M parameters to accomplish this and introduce the \textbf{Co}ordinated \textbf{M}asking and \textbf{W}avelet \textbf{E}nhancement  {(CoMWE)} module. It consists of two key components: \textbf{R}e-\textbf{M}asked \textbf{K}nowledge \textbf{D}istillation {(RMKD)}, which encodes richer content knowledge by applying strategically reconstructed masks with a high mask ratio; \textbf{D}etail-\textbf{A}ware \textbf{W}avelet  \textbf{E}hancement  {(DAWE)}, which preserves fine-grained details by leveraging high-frequency wavelet components.
In the second stage, \textbf{Full-Disc Adaptation}, we present \textbf{Self-MaskTune}, a simple yet effective fine-tuning method that generalizes precipitation retrieval from scanning swath to full-disc. Self-MaskTune achieves this by balancing multimodal knowledge and full-disc IR knowledge through guided parameter updates.
This pipeline enables PRE-Net to effectively learn from multimodal data while addressing the constraints of narrow coverage and missing modalities, ultimately enhancing full-disc precipitation retrieval.

Our contributions are as follows:
(i) We introduce the Precipitation Retrieval Expansion task, aiming at high-quality IR-based full-disc precipitation retrievals, and provide a benchmark.
(ii) We propose PRE-Net, featuring a multimodal knowledge expansion pipeline with two stages: Swath-Distilling and Full-Disc Adaptation, enabling effective multimodal knowledge transfer to overcome quality and coverage difference limitations.
(iii) We introduce the CoMWE module to effectively encode richer knowledge and high-frequency components, and Self-MaskTune to adaptively balance knowledge distilled from multimodal data with new knowledge learned from full-disc IR.
(iv) Experimental results demonstrate that PRE-Net surpasses several IR-based precipitation products (i.e., PERSIANN-CCS, PDIR) and achieves performance comparable to the multimodal-based algorithm IMERG.

\section{Related Work}
\subsection{Precipitation Retrieval}
\subsubsection{Traditional Precipitation Retrieval}
Numerous precipitation retrieval algorithms based on satellite observations have been proposed. Early approaches relied on physical models and empirical relationships. For instance, IR-based algorithms, such as the GPI\cite{5} and the GMSRA \cite{6}, model the relationship between BTs and precipitation. Similarly, the GPROF \cite{12} used Bayesian methods to match PMW with rainfall profiles. With the advent of machine learning, data-driven approaches\cite{18, pmwdeeplearn2, ir1} have gained prominence in precipitation retrieval. Among these, PERSIANN\cite{7} and its derivatives (e.g., PERSIANN-CCS~\cite{8} and PDIR~\cite{9}), which are Satellite-based Precipitation Products (SPPs) based on IR data, are widely used by the scientific community.

\subsubsection{Multimodal Precipitation Retrieval}
Efforts to integrate data from multiple sensors have produced algorithms like Deep-STEP \cite{gorooh2022deep} and others \cite{20, 22}. However, assuming complete multimodal data is often unrealistic due to the sparse and asynchronous nature of satellite observations. 
Some physical-based methods expand the swath observations by calculating cloud motion vectors to deal with modality sparsity~\cite{14, 15, panegrossi1998use}.
The Integrated Multi-satellite Retrievals for GPM (IMERG)~\cite{14} is the most authoritative SPP for precipitation monitoring and research. IMERG integrates multimodal for bias correction, providing near-global precipitation estimates at a high spatiotemporal resolution (0.1° × 0.1°, 30-minute intervals). 
Whereas, these physical methods require substantial computing resources, and exhibit systematic biases in complex topographies and seasonality \cite{23, 24}. Moreover, IMERG suffers from limited real-time availability. Even the products designed for near-real-time usage have latencies of 4–14 hours.
Additionally, attempts to expand swath precipitation using statistical methods \cite{19} have been made, but these approaches often lead to increasing errors as the expansion scope grows. Therefore, we propose the Precipitation Retrieval Expansion (PRE) task to achieve high-quality, high-resolution, and low-cost precipitation retrievals with global coverage.

\begin{figure*}[t]
\centering
\includegraphics[width=0.98\textwidth]{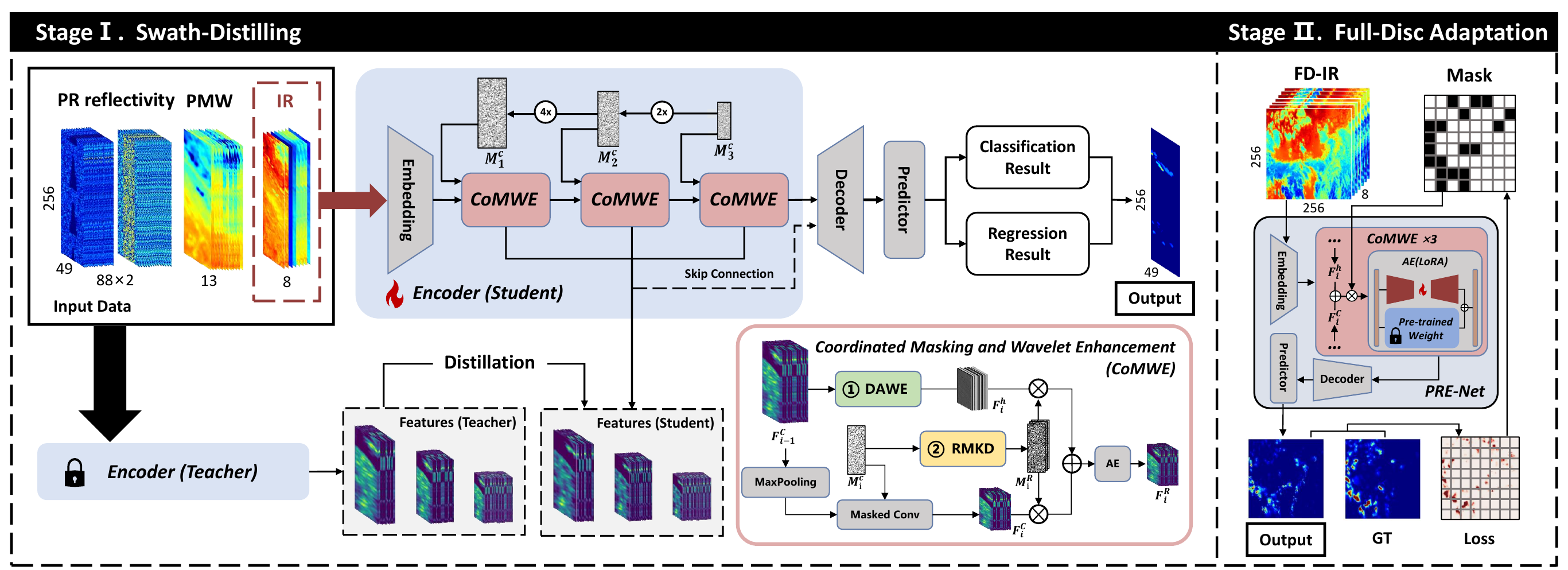} 
\caption{The multimodal knowledge expansion pipeline of PRE-Net. The framework of PRE-Net is detailed in Stage I. 
Two identical PRE-Net models are trained to obtain both classification and regression results.
}
\label{fig:fig2}
\end{figure*}

\subsection{Incomplete Multimodal Learning} 
Most multimodal learning tasks assume access to complete data, but this is unrealistic due to device limitations, privacy concerns, etc. To address incomplete multimodal learning, recent efforts focus on two main approaches: generative-based and learning-based methods. 
Generative-based methods attempt to recreate missing modalities, converting tasks into complete multimodal tasks \cite{hgan, cycae, smil}. These methods are effective when sufficient paired multimodal data is available for training.
Learning-based methods focus on learning representations from complete multimodal. These methods can be further divided into modality-agnostic learning and knowledge distillation (KD). 
Modality-agnostic learning methods\cite{mmformer, hemis, hetore} capture modality-invariant features adaptable to any missing modality scenario. These methods are well-suited where the missing patterns are diverse but the correlation strength between modalities and targets is balanced.
KD transfers knowledge from a teacher model (trained on complete modalities) to a student model (handling incomplete inputs). KD can be categorized into two types: response-based KD \cite{prototype, mlim}, which utilizes soft labels for distillation, and feature-based KD \cite{mmanet, disentangle}, which leverages intermediate features to capture richer knowledge. In this work, we adopt feature-based KD for PRE as it transfers precipitation-related knowledge from strongly correlated PMW/PR to IR.

\section{Methodology}
\subsection{Task Definition}

Formally, we define $X =\{X_{IR}, X_{PMW}, X_{PR}\}$ as the multi-modality data set consisting of infrared, passive microwave, and spaceborne radar reflectivity, where $X_{IR}$ is available in both swath and full-disc coverage, while $X_{PMW}$ and $X_{PR}$ are limited to swath coverage. The corresponding precipitation labels are denoted by $Y = \{Y^s, Y^f\}$, representing the precipitation ground truth over the swath and full-disc areas. During training, the integrated multi-modal data $X$ is used to predict precipitation $Y$. At inference time, given only an infrared observation $x_{IR}\in X_{IR}$, the model predicts the full-disc precipitation ${\hat{y}}^f$, thereby enabling high-quality precipitation retrieval beyond the swath coverage.
PRE task includes two targets: classification of precipitation pixels and regression of precipitation intensity. 

\subsection{Overview of PRE-Net and Multimodal Knowledge Expansion}
The Multimodal Knowledge Expansion (MKE) pipeline is illustrated in Fig.\ref{fig:fig2}.
PRE-Net is a U-shape model that incorporates specialized Coordinated Masking and Wavelet Enhancement (CoMWE) modules on the encoder side, 
which enables PRE-Net to encode richer and more detailed features. 
Overall, the MKE pipeline consists of the Swath-Distilling and Full-Disc Adaptation stages. In Stage I, we pre-train a multimodal UNet model that integrates various swath data, including $X_{IR}, X_ {PMW}, X_ {PR}$, to learn joint feature representations for precipitation retrieval. This pre-trained model then serves as a multimodal teacher, transferring its knowledge to a single-modal PRE-Net, which performs precipitation retrieval using only swath IR data.
In Stage II, the single-modal PRE-Net undergoes Full-Disc Adaptation, where we employ LoRA~\cite{lora} and Self-MaskTune to refine its ability to perform full-disc precipitation retrieval while keeping the knowledge transferred from the teacher model to handle the broader spatial coverage provided by IR data.
The two-stage pipeline enables PRE-Net to bridge the gap between multimodal data integration and effective single-modality generalization, enhancing the accuracy and robustness of precipitation retrieval.
As a result, PRE-Net can achieve high-quality full-disc retrieval relying solely on IR data.

\subsection{Swath-Distilling: Coordinated Masking and Wavelet Enhancement}
We first integrate multimodal data (IR, PMW, and PR) and train a teacher UNet model for scanning swath precipitation retrieval. This multimodal fusion significantly outperforms single-modality baselines and performs better than GPM 2B-CMB\cite{TheGPMCombinedAlgorithm}, a leading satellite precipitation product (SPP) (see Supplementary \ref{teacher_details} for architecture and ablation studies).  
To generalize this capability from swath-scale to full-disc precipitation retrieval, we employ feature-based knowledge distillation, where a student PRE-Net model, restricted to IR-only inputs, aligns its feature maps with the teacher’s multimodal representations.

However, the teacher’s multimodal inputs and the student’s IR-only inputs create a significant feature representation gap.
Conventional feature-based KD struggles to bridge this asymmetry, as the student lacks access to the teacher’s complementary PMW/PR data.
To enhance feature learning during KD, we leverage Masked Image Modeling (MIM). MIM masks random input patches and trains models to reconstruct missing features from visible context.
A well-known advancement in MIM is Masked Autoencoder (MAE)~\cite{mae}.
While MAE demonstrates exceptional performance at high mask ratios (e.g., 75\% masking), its integration in the Masked Knowledge Distillation (MKD)~\cite{mkd} reveals a critical mismatch: MKD achieves optimal distillation effectiveness at low mask ratios (~ 20\% masking), failing to exploit MAE’s high-masking reconstruction strengths. 
This mismatch between MAE’s high-masking design and MKD’s low-masking operational regime creates inefficiencies, as MKD fails to fully exploit MAE’s reconstruction capabilities under its intended high-masking paradigm. This inefficiency limits the student’s ability to recover high-quality precipitation patterns from partial multimodal observations (i.e., IR only).


To resolve this contradiction, we propose the Coordinated Masking and Wavelet Enhancement (CoMWE) module, which adaptively balances high-ratio masking with detail preservation. As illustrated in Fig.~\ref{fig:fig2}, CoMWE combines two components: RMKD and DAWE. The process of CoMWE can be expressed as follows:


\begin{equation}
    F_i^c=MaskedConv(\phi(F_{i-1}^c)\times M_i^c;\theta_{MC}),
    \label{eq:eq1}
\end{equation}
\begin{equation}
    F^h_i = DAWE(embed(X_{IR})),
    \label{eq:eq2}
\end{equation}
\begin{equation}
    F_i^R=AE((F_i^c + F^h_i)\times M_i^R;\theta_{AE}),
    \label{eq:eq3}
\end{equation}
where $MaskedConv$ and $AE$ represent multi-scale masked convolutions and auto-encoders. $\theta_{MC}$ and $\theta_{AE}$ are the corresponding weights. $\phi$ is a max-pooling and convolutional layer with kernel size 1 × 1 to align the down-sampling channel dimension. $F_i^c$, $F^h_i$ and $F_i^R$ represent multi-scale convolution features, high-frequency prompt features, and reconstructed features, respectively. $M_i^c, M_i^R$ are masks for multi-scale masked convolutions and auto-encoders. Eq.\ref{eq:eq2} will be described in the following part.

\begin{figure}
    \centering
    \includegraphics[width=0.98\linewidth]{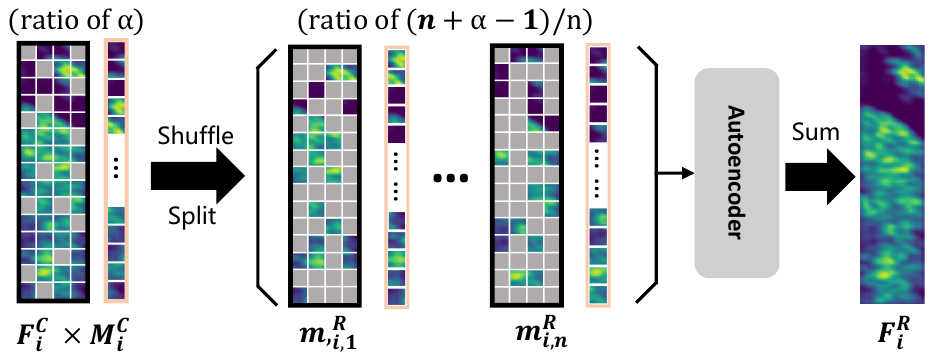}
    \caption{The illustration of RMKD. The inputs are the feature map $F_i^c$ and the mask $M_i^c$.}
    \label{fig:fig3}
\end{figure}

\subsubsection{Re-Masked Knowledge Distillation.}
We develop the RMKD, which incorporates different masks tailored for $F_i^C$ to reconstruct $F_i^R$.
As illustrated in Fig.\ref{fig:fig3}, $M_i^c$ with a predefined mask ratio $\alpha$ is applied for the multi-scale convolutional layers. This initial mask $M_i^c$ conceals few features, enabling the model to learn robust representations from visible patches. To optimize feature reconstruction with high mask ratios, the remaining visible patches from $F_i^c$ are further shuffled and equally partitioned into $N$ parts. Then, these $N$ parts are combined with a mask set $M_i^R = \{m_{i,1}^R, m_{i,2}^R, \ldots, m_{i,n}^R\}$ separately, resulting in $N$ mask patches, each with a high mask ratio of $\frac{n + \alpha - 1}{n}$. These masks are then independently applied to $F_i^c$ and $F_i^h$, which are fed into an Auto-Encoder for feature reconstructions. Finally, $F_i^R$ combines reconstructed features with learnable weights.

\begin{figure}
    \centering
    \includegraphics[width=0.98\linewidth]{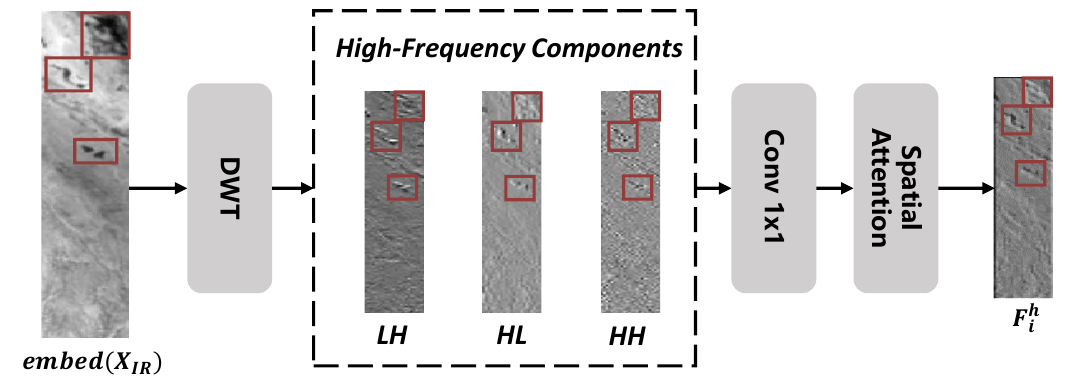}
    \caption{The illustration of DAWE. It decomposes input images or feature maps into wavelet coefficient components. The red highlighted areas are high-frequency details of precipitation-related features. DAWE precisely preserves these fine-grained features and reconstructs them with sharp boundaries.}
    \label{fig:fig4}
\end{figure}

\subsubsection{Detail-Aware Wavelet Enhancement.}
High-frequency details such as precipitation boundaries and local textures, are critical in accurate knowledge transfer during distillation. However, increasing the mask rate could amplify erroneous high-frequency components (e.g., spurious edges or noise)~\cite{how_mask}, while conventional masking combined with pooling layers further degrades high-frequency components.
To address this issue, we propose the Detail-Aware Wavelet Enhancement (DAWE) module (Fig.\ref{fig:fig4}), which leverages Discrete Wavelet Transform (DWT) to preserve and enhance high-frequency details, ensuring more accurate feature recovery.

DWT decomposes input images or feature maps into four subbands: Low-frequency (LL): captures global semantic patterns (e.g., large-scale precipitation systems).
High-frequency (HL, LH, HH): encodes directional details—horizontal (HL), vertical (LH), and diagonal (HH) components critical (e.g. fine-grained precipitation boundaries)~\cite{dwt1192}.
While the high-frequency subbands inherently contain noise, they retain essential structural details. 
DAWE first extracts and concatenates the HL, LH, and HH components. A $1\times1$ convolutional layer then aligns their channels. To suppress noise while enhancing discriminative details, a spatial attention (SA) mechanism is applied, which dynamically recalibrates the weights of high-frequency features based on their spatial relevance to precipitation patterns. 
The process of DAWE can be formulated as follows:
\begin{equation}
    LL_i, (HL_i, LH_i, HH_i) = DWT(embed(X_{IR}); f_{haar}),
    \label{eq:eq4}
\end{equation}
\begin{equation}
    F_i^h = SA(\phi(cat(HL_i, LH_i, HH_i))),
    \label{eq:eq5}
\end{equation}
where $\phi$ denotes a $1\times1$ convolutional layer for channel alignment, $SA$ is the Spatial Attention module. Eq.\ref{eq:eq4} denotes performing $i$-th DWT on $embed(X_{IR})$ with the Haar wavelet filter.

\subsubsection{Optimization Target.}
PRE-Net concludes with a lightweight predictor module, which transforms the decoder’s output features into precipitation categories or estimates. 
This predictor employs a $1\times1$ convolutional layer to project the feature maps into a single-channel representation, followed by a sigmoid function for precipitation classification.
The training target for PRE-Net consists of three losses:
(i) task loss $\mathcal{L}_{task}$, e.g., cross-entropy (CE) loss for classification and mean-squared error (MSE) loss for regression; (ii) feature-based knowledge distillation loss $\mathcal{L}_{feat}$ using KL divergence to align the latent space of the student with the teacher's, and (iii) feature-reconstruction loss $\mathcal{L}_{rec}$, enforcing pixel-level consistency between the student’s reconstructed features from masked inputs and the teacher’s original features.
The overall training loss in the swath-distilling stage is as follows:
\begin{equation}
    \mathcal{L}_{SD}=\mathcal{L}_{task}+\lambda\mathcal{L}_{feat}+\gamma\mathcal{L}_{rec},
\end{equation}
where $\lambda$ and $\gamma$ are parameters to balance the distillation losses. 

\subsection{Full-Disc Adaptation: Self-MaskTune}
In the Full-Disc Adaptation stage, the model transitions from scanning swath to full-disc precipitation retrieval. A core challenge is maintaining the multimodal knowledge distilled in Stage I while adapting to the distributional characteristics of full-disc IR data. A naive strategy, i.e., freezing the CoMWE module and fine-tuning only the decoder, yields suboptimal results, as it cannot fully account for the differences between swath and full-disc precipitation patterns. Conversely, allowing CoMWE to be entirely learnable risks knowledge forgetting, since IR-only data, though abundant, lacks the reliability of multimodal inputs for precise retrieval. To address this, we adopt a hybrid strategy: LoRA enables efficient adaptation to new IR-based knowledge, while Self-MaskTune preserves previously acquired multimodal knowledge by selectively updating model parameters to account for high knowledge disparity.

To avoid severe knowledge forgetting of the CoMWE module during fine-tuning, we selectively update only the auto-encoder (AE) parameters. While convolutional layers transform single-modal inputs into multimodal-compatible features, the AE is tasked with refining these features for precipitation retrieval. Freezing the convolutional layers preserves their pretrained multimodal mapping capabilities, whereas adapting the AE ensures targeted refinement of task-specific features.
Specifically, we incorporate the Low-Rank Adaptation (LoRA)\cite{lora} into the AE, constraining parameter updates via bypass learning: $W_{updated} = W_{pretrained} +\Delta W$. To guide the finetuning process, we propose Self-MaskTune (Fig.\ref{fig:fig2}), a novel dynamic masking strategy that prioritizes regions with high prediction uncertainty, where swath-based multimodal knowledge conflicts with full-disc IR patterns. By dynamically prioritizing these regions, the model generalizes beyond spatially confined precipitation systems (e.g., convective cores), reducing overfitting to swath-specific biases.
Specifically, during the first K epochs, no masking is applied, allowing the model to adapt to full-disc IR characteristics. After K epochs, mask $M^{AE}(y, {t-1})$ is introduced to highlight regions in $F^R$ with high prediction errors in $\mathcal{L}_{task}$, which requires relearning to improve generalization:
\begin{equation}
    F^R =
    \begin{cases}
    AE((F^c + F^h);\theta_{LoRA}), &t \leq K,\\
    AE((F^c + F^h)\times M^{AE}(y^f,t-1);\theta_{LoRA}) , &t > K.
    \end{cases}
\end{equation}

\noindent $M^{AE}(y^f, {t-1})$ is a mask that identifies regions of sample $y$ with higher prediction errors in the previous $t-1$ epoch, and is applied to the feature map within AE. To obtain this mask, we introduce $\mathcal{P}_\text{max}$ as a threshold for determining whether a feature region requires further learning:
\begin{equation}
    \mathcal{P}_\text{max}=\rho\cdot\text{max}(\mathcal{L}_{task}^{t-1}),
\end{equation}
\begin{equation}
    M^{AE}(y^f,t-1)=
    \begin{cases}
    1, &\text{where}\ \mathcal{L}^{t-1}_{task}(y^f, \hat{y}^f)\geq\mathcal{P}_\text{max},\\
    0, &\text{otherwise}.
    \end{cases}
\end{equation}
where $y^f$ and $\hat{y}^f$ denote the ground truth and prediction for full-disc precipitation, respectively. $\rho$ is the parameter to adjust the threshold. The training loss in the Full-Disc Adaptation stage is formulated as:
\begin{equation}
\mathcal{L}_{FF}={\mathcal{L}_{task}(y^f,\hat{y}^f)}, \text{subject to}\  M^{AE}(y^f, {t-1}).
\end{equation}
With this loss, Self-MaskTune first consolidates prior knowledge and assimilates new knowledge, then incrementally incorporates new knowledge with uncertainness guidance, achieving a balance between the two kinds of knowledge.

\begin{table}[t]
\centering
\small 
\setlength{\tabcolsep}{0.5mm}
\renewcommand{\arraystretch}{1.6}
{%
\begin{tabular}{ccccc}
\toprule
  \begin{tabular}[c]{@{}c@{}}Data Type\end{tabular} &
  \begin{tabular}[c]{@{}c@{}}Time \\ Frequence\end{tabular} &
  \begin{tabular}[c]{@{}c@{}}Spatial \\ Resolution\end{tabular} &
  \begin{tabular}[c]{@{}c@{}}Spatial \\ Coverage\end{tabular} & 
  \begin{tabular}[c]{@{}c@{}}Data \\ Usage\end{tabular} 
  \\
  \midrule
  IR (FY4A/AGRI) & 0.07 h & 4 km & Full-disc & Train / Infer \\
  PMW (GPM/GMI) & 1.5 h & 13 km & Swath & Train \\
  PR (GPM/DPR) & 1.5 h & 5 km & Swath & Train \\
  \midrule
  CLDAS-V2.0 & 1 h & $\sim$7 km & Full-disc & Ground Truth \\
  \midrule
  PERSIANN-CCS & 1 h & $\sim$4.5 km & Full-disc & \multirow{3}{*}{Competitive} \\
  PDIR & 1 h & $\sim$4.5 km & Full-disc &  \\
  GPM 2B-CMB & 1 h & 5 km & Swath & Products  \\
  IMERG & 0.5 h & $\sim$11 km & Full-disc &  \\
  \bottomrule
\end{tabular}%
}
\caption{Data and products used in this study.}
\vspace{-2em}
\label{tab:tab1}
\end{table}

\section{Experimental Setups}
\subsection{Benchmark}
\subsubsection{Study Area and Data Collection}
This study focuses on the retrieval of precipitation over the region spanning 0–65°N and 60–160°E, covering East Asia and adjacent areas. 

As shown in Table \ref{tab:tab1}, the data used in this work include multimodal observations with varying temporal and spatial resolutions. IR data from FY-4A/AGRI serves as the primary input for both training and inference, featuring a temporal resolution of 0.08 hours and a spatial resolution of 4 km for full-disc coverage. Passive microwave (PMW) data from GPM/GMI and precipitation radar (PR) data from GPM/DPR provide cloud microphysical properties, offering scanning swath coverage with a temporal resolution of 1.5 hours and respective spatial resolutions of 13 km and 5 km.

\begin{table}[t]
    \vspace{-1em}
    \centering
    \small 
    \setlength{\tabcolsep}{1.1mm}
    \renewcommand{\arraystretch}{1.5}{%
    \begin{tabular}{@{\hspace{0mm}}cc|ccccc@{\hspace{0mm}}}
        \toprule
        \textbf{RMKD} & \textbf{DAWE} & \textbf{RMSE $\downarrow$} & \textbf{CC $\uparrow$} & \textbf{POD $\uparrow$} & \textbf{FAR $\downarrow$} & \textbf{CSI $\uparrow$} \\
        \midrule
        $\times$ & $\times$ & 0.9782 & 0.3408 & 0.2159 & 0.4536 & 0.1831 \\
        $\checkmark$ & $\times$ & 0.9484 & 0.4051 & 0.4288 & 0.4702 & 0.3106 \\
        $\times$ & $\checkmark$ & 0.9689 & 0.3859 & 0.2456 & \textbf{0.3772} & 0.2138 \\
        $\checkmark$ & $\checkmark$ & \textbf{0.9457} & \textbf{0.4106} & \textbf{0.4921} & 0.4944 & \textbf{0.3322} \\
        \bottomrule
    \end{tabular}%
    }
    \caption{Ablations of CoMWE in the Swath-Distilling stage, with the best results in bold.}
    \vspace{-3em}
    \label{tab:AbCoMWE}
\end{table}

\subsubsection{Datasets Construction}
CLDAS-V2.0~\cite{shi2014status} is used as the ground truth for evaluation due to its superior performance over other products when validated on ground station observations in the study region. Competitive products, PERSIANN-CCS\cite{8}, PDIR\cite{9}, GPM 2B-CMB\cite{TheGPMCombinedAlgorithm}, and IMERG\cite{14}, are Numerical Weather Prediction (NWP) methods~\cite{bauer2015quiet} used for comparison.

\textbf{Our datasets consist of two splits: Swath-MPR for the Swath-Distilling stage and Full-IPR for the Full-Disc Adaptation stage as well as the PRE task.} We evaluate the model's overall performance across both swath and full-disc scenarios.

\noindent \textit{Swath-MPR}:
We collected multimodal data from 2021 to 2022 as the Swath multimodal precipitation retrieval dataset (Swath-MPR) for training, which includes IR, PMW, PR reflectivity, geographic information, and precipitation products: CLDAS-V2.0 and GPM 2B-CMB. PMW and PR observations are sourced from the GPM Core Observatory (GPM-CO), which provides precise timestamps for each swath entry into the study region. For each PMW/PR swath, we identify the closest FY-4A IR data and CLDAS-V2.0 in time, ensuring a maximum temporal deviation of 10 minutes to minimize mismatches. Spatially, all data were spatially matched at a resolution of 5 km and processed with a crop $256\times 49$. We excluded swaths with minimal or no precipitation, as they contribute little to the variance of assessment metrics \cite{25}. Swath-MPR consists of 3,597 swaths, of which 2281 swaths and 279 swaths from 2021 are used for training and validation, and 1037 swaths from 2022 for testing.

\noindent \textit{Full-IPR}:
We also collected IR and precipitation data from CLDAS-V2.0 for 2022 and 2023 to create the full-disc IR precipitation retrieval dataset (Full-IPR) for finetuning and evaluating the PRE-Net. We aligned these data at a spatial resolution of 4 km and performed patch-clipping with a $256\times 256$ crop. Patches with minimal or no precipitation were excluded. The 2022 data were split into a training set (first 10 months) and a validation set (last two months), with all 2023 data used for testing. In total, 6,220 patches were used for training, 754 patches for validation, and 12,819 patches for testing.

\subsubsection{Evaluation Metrics}
We adopt a set of statistical metrics to quantify the performance of the various methods. These metrics are selected based on their established effectiveness in assessing precipitation retrievals, as demonstrated in prior studies\cite{ir1,19,ir2}.

For precipitation classification, we use the Probability of Detection (POD), False Alarm Rate (FAR), and Critical Success Index (CSI), with CSI acknowledged as the most balanced metric. A precipitation threshold of 0.1 mm/hr is applied to distinguish rainy and non-rainy pixels, consistent with standard practices. Besides, to address resolution discrepancies between comparison products (PERSIANN-CCS and PDIR at 0.04°, IMERG at 0.1°) and PRE-Net's results (at 4 km), we introduce CSI-Neighbor (CSI-4 and CSI-8)\cite{25}. These metrics, calculated using max pooling with kernel sizes of 4 and 8, assess local spatial consistency between predicted and observed precipitation patterns.

For precipitation intensity estimation, we utilize the Root Mean Square Error (RMSE) and Pearson Correlation Coefficient (CC). RMSE quantifies the magnitude of error between the retrieved precipitation and the GT, while CC measures the linear relationship between the predicted and GT intensities. 


\subsection{Implementation Details}
We implemented PRE-Net with PyTorch. Both training and inference are conducted on NVIDIA RTX A6000 GPUs. We used RMSprop to optimize PRE-Net with an initial learning rate of 1e-6. We decrease the learning rate by a factor of 0.5 every 30 epochs. The training batch size is set at 32, and the epoch is set at 200, and 100 in two stages respectively. The initial mask ratio $\alpha$ is set at 0.25, and the number of generated masks $n$ in RMKD is set to 3. The $\lambda$ is set to 0.2, $\gamma$ is set to 50. We selected the model that performed best on the validation set and reported its performance on the test set. 

\section{Experimental Results}

\begin{table}[t]
    \vspace{-1em}
    \centering
    \small 
    \setlength{\tabcolsep}{0.6mm} 
    \renewcommand{\arraystretch}{2} 
    \begin{tabular}{@{\hspace{0mm}}c|ccccc@{\hspace{0mm}}}
        \toprule
         \textbf{Method} & \textbf{RMSE $\downarrow$} & \textbf{CC $\uparrow$} & \textbf{POD $\uparrow$} & \textbf{FAR $\downarrow$} & \textbf{CSI $\uparrow$} \\
        \midrule
        UNet & 0.9782 & 0.3408 & 0.2159 & 0.4536 & 0.1831 \\
        \midrule
        KD~\cite{kd} & 0.9807 & 0.3567 & 0.2927 & \textbf{0.4323} & 0.2393 \\
        MGD~\cite{mgd} & 0.9779 & 0.3429 & 0.3036 & 0.4720 & 0.2388 \\
        MKD~\cite{mkd} & 0.9541 & 0.3940 & 0.3654 & 0.4430 & 0.2831 \\
        \midrule
        RMKD 
        & \textbf{0.9484} & \textbf{0.4051} & \textbf{0.4288} & 0.4702 & \textbf{0.3106} \\
        \bottomrule
    \end{tabular}
    \caption{Ablations of knowledge distilling methods in the Swath-Distillation stage, with the best results in bold.}
    \vspace{-3em}
    \label{tab:kd_methods}
\end{table}

\subsection{Ablation Studies}
To validate the effectiveness of our MKE pipeline and module designs, we conduct ablation studies across three key dimensions: (i) CoMWE module components, (ii) knowledge distillation methods in the Swath-Distilling stage, and (iii) finetuning strategies in the Full-Disc Adaptation stage.
\begin{table*}[t]
    \vspace{-1em}
    \centering
    \small 
    \setlength{\tabcolsep}{3.0mm}
    \renewcommand{\arraystretch}{1.5}{%
    \begin{tabular}{@{\hspace{2mm}}c|c|c|ccccc@{\hspace{2mm}}}
        \toprule
        \multicolumn{3}{c|}{\textbf{Method}}
        & \textbf{RMSE $\downarrow$} & \textbf{CC $\uparrow$} & \textbf{POD $\uparrow$} & \textbf{FAR $\downarrow$} & \textbf{CSI $\uparrow$} \\
        \midrule
        Train from scratch& \multicolumn{2}{|c|}{-} & 0.9082 & 0.4589 & 0.3309 & 0.3169 & 0.2869 \\
    
        \midrule
        \multirow{5}{*}{Train with MKE}
        & KD & Non-MaskTune & 0.8976 & 0.4613 & 0.3649  & 0.3357 & 0.3081 \\
        \cmidrule(lr){2-8}
        & \multirow{4}{*}{CoMWE}
        & / & 1.0291 & 0.1063 & 0.1173 & 0.4606 & 0.1066 \\
        & & Non-MaskTune & 0.8990 & 0.4616 & 0.3171 & \textbf{0.3006} & 0.2791 \\
        & & Rand-MaskTune & 0.8853 & 0.4850 & 0.4136 & 0.3887 & 0.3274\\
        & & Self-MaskTune & \textbf{0.8714} & \textbf{0.5031} & \textbf{0.5598} & 0.4349 & \textbf{0.3912} \\
        \bottomrule
    \end{tabular}%
    }
    \caption{Ablations of strategy combinations for full-disc precipitation retrieval, with the best results in bold.}
    \vspace{-2em}
    \label{tab:Ab2}
\end{table*}

\noindent \textbf{\textit{CoMWE components.}} 
The ablation results for the CoMWE module presented in Table \ref{tab:AbCoMWE} validate its design. The results are reported on the Swath-MPR test set. First, only RMKD achieves significant improvements over the baseline, particularly in CC(+5.57\%), POD (+21.29\%), and CSI (+12.75\%). This demonstrates that RMKD effectively enhances the feature representation by strategically masking less relevant regions and focusing on content-rich knowledge. 
Second, DAWE alone leads to suboptimal performance. While it reduces false alarms with FAR being 0.3772, it is inferior to RMKD alone in other metrics because it amplifies noise while enhancing detailed knowledge.
Finally, combining RMKD and DAWE achieves optimal results (CC: 0.4106, CSI: 0.3322). 
These results validate the design of the CoMWE module, demonstrating that RMKD and DAWE are complementary and their integration is essential for achieving optimal precipitation retrieval performance.

\begin{figure}[t]
    \centering
    \includegraphics[width=0.98\linewidth]{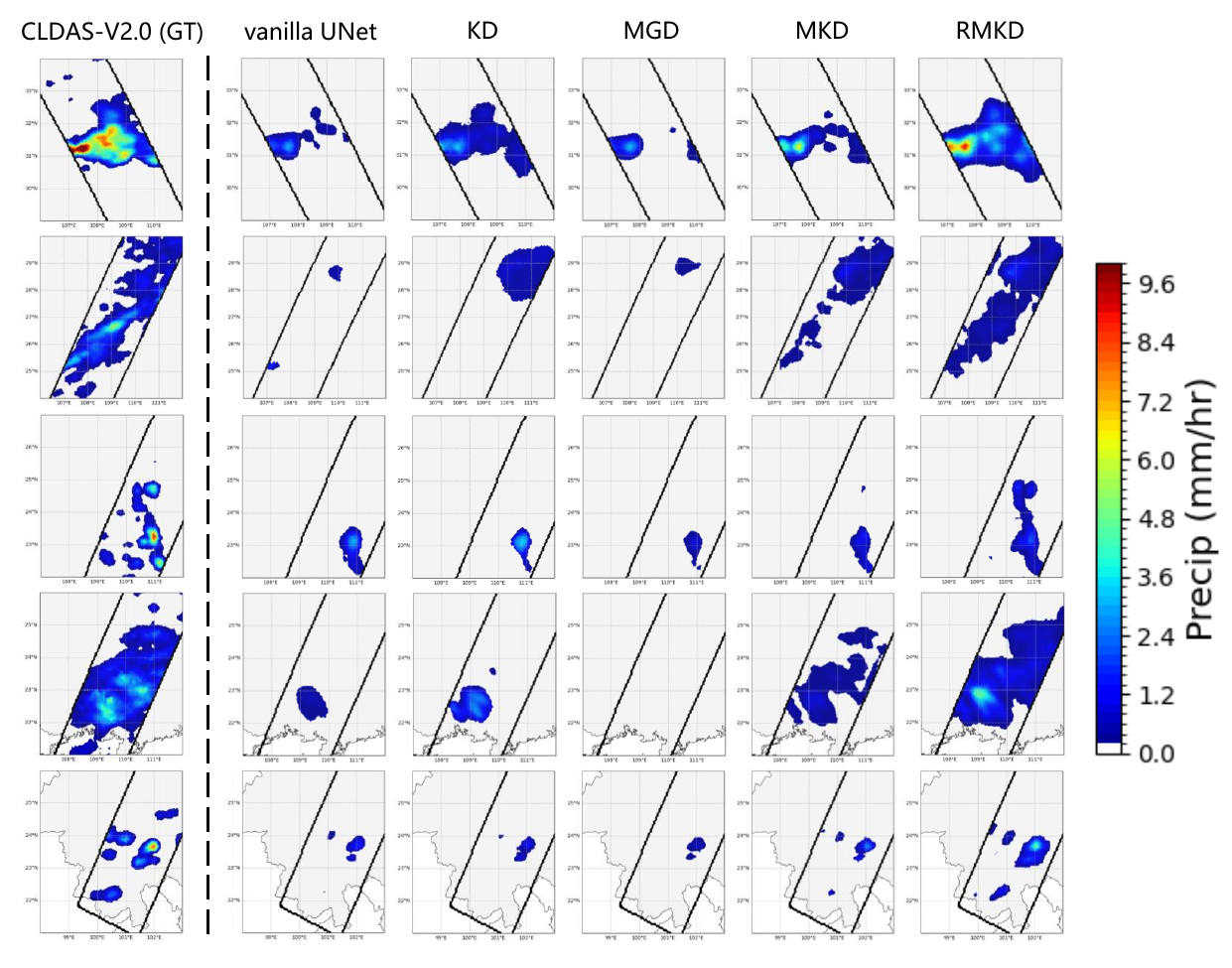}
    \vspace{-1em}
    \caption{Visual comparison on diverse scanning swaths.}
    \vspace{-2em}
    \label{fig:fig6}
\end{figure}

\noindent \textbf{\textit{Distilling strategies.}} 
Table \ref{tab:kd_methods} compares various knowledge distillation strategies applied during the Swath-Distilling stage. The results are reported on the Swath-MPR test set. The UNet baseline, trained solely on IR data, achieves limited performance (CC: 0.3408, and CSI: 0.1831), highlighting the challenge of IR-only precipitation retrieval.
Incorporating vanilla knowledge distillation results in modest improvements ( CC  +1.59\%, and CSI +5.62\%), confirming the value of distilling multimodal teacher knowledge into the IR-based student.
MGD, by simply introducing a masking strategy without refining the multimodal features through AE, does not demonstrate better performance compared to vanilla KD.
MKD further boosts performance (CC +5.32\%, and CSI +10.0\%) by combining masked feature reconstruction with distillation.
Our RMKD method achieves the best results (CC: 0.4051, CSI: 0.3106), 
with 12.75\% CSI improvement over baseline, demonstrating its ability to balance masking ratio and feature reconstruction for representation learning.
In addition, RMKD significantly increases POD (+21.29\%), and decreases RMSE to 0.9484 (-3.08\%), underscoring its ability to transfer multimodal knowledge to the IR-only student.

The visualization in Fig.\ref{fig:fig6} compares different KD methods for Swath-Distilling. The UNet baseline, KD, and MGD underestimate precipitation intensity and coverage. MKD slightly enhances precipitation details, but it still fails in some areas. RMKD recovers the most accurate and detailed precipitation patterns, closely matching the ground truth in both intensity and spatial distribution. 

\begin{table*}[t]
    \vspace{-0.5em}   
    \centering
    \small 
    \setlength{\tabcolsep}{3.0mm} 
    \renewcommand{\arraystretch}{1.5} 
    \begin{tabular}{@{}c|ccccccc@{}}
        \toprule
        \textbf{Method} & \textbf{RMSE $\downarrow$} & \textbf{CC $\uparrow$} & \textbf{POD $\uparrow$} & \textbf{FAR $\downarrow$} & \textbf{CSI $\uparrow$} & \textbf{CSI-4 $\uparrow$} & \textbf{CSI-8 $\uparrow$} \\
        \midrule
        PERSIANN-CCS & 1.6058 & 0.2147 & 0.2851 & 0.5230 & 0.2172  & 0.2928 & 0.3672 \\
        PDIR         & 4.2461 & 0.0833 & 0.4587 & 0.5861 & 0.2781  & 0.3573 & 0.4354 \\
        IMERG & 1.1074 & 0.4484 & 0.5109 & 0.4777 & 0.3482 & 0.4176  & \textbf{0.4891} \\
        \midrule
        PRE-Net      & \textbf{0.8714} & \textbf{0.5028} & \textbf{0.5597} & \textbf{0.4349} & \textbf{0.3912}  & \textbf{0.4405} & 0.4815 \\
        \bottomrule
    \end{tabular}
    \caption{Comparison with precipitation retrieval products. The PERSIANN-CCS, PDIR, and IMERG are NWP methods, IMERG is a multimodal-based retrieval method, while the others are IR-based methods.}
    \vspace{-2em}
    \label{tab:comparison_methods}
\end{table*}

\begin{figure*}[t!]
    \centering
    \includegraphics[width=0.98\textwidth]{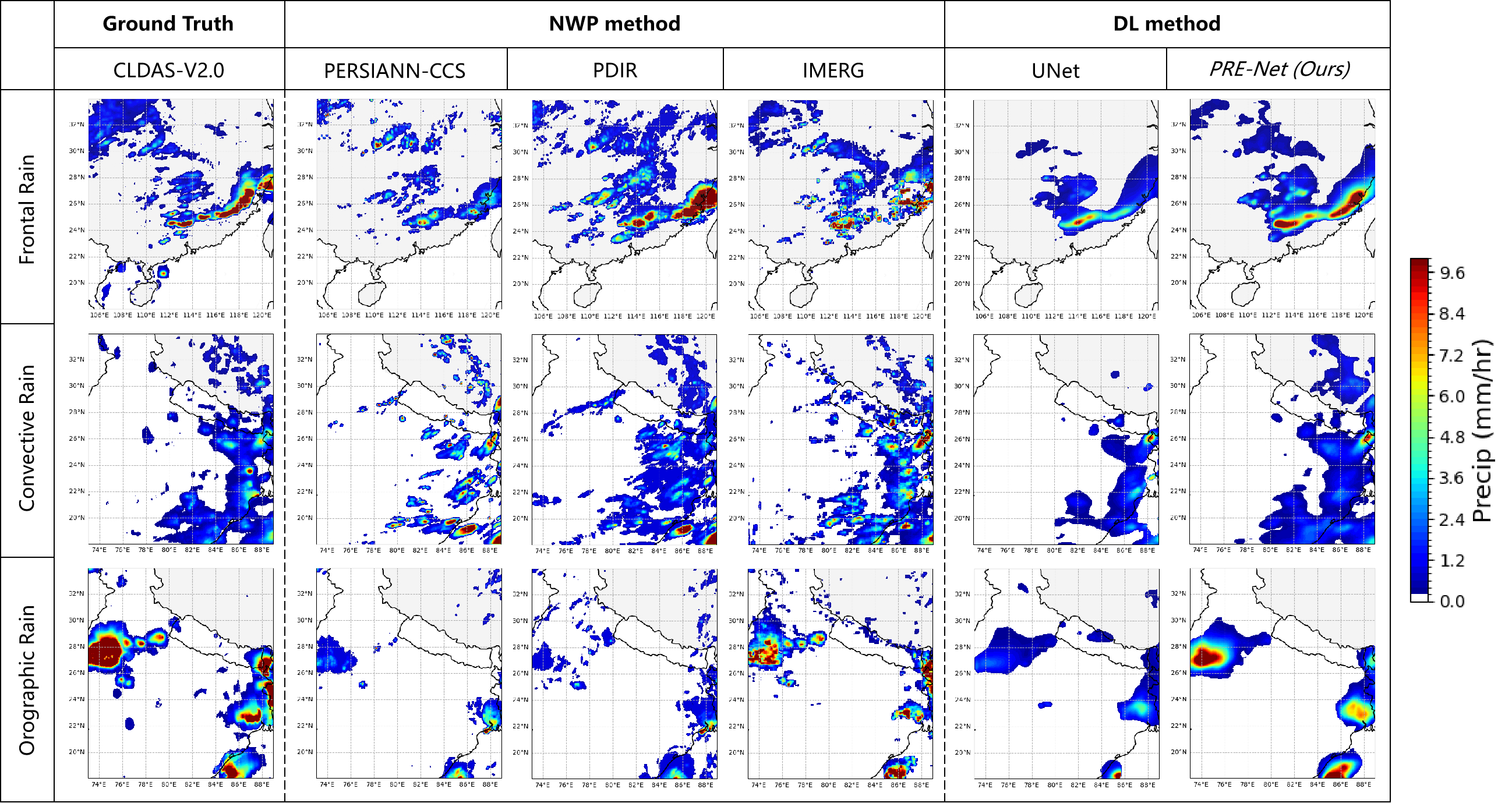}
    \vspace{-0.5em}
    \caption{The visual comparison examples on different full-disc precipitation events.}
    \vspace{-1.5em}
    \label{fig:fig5}
\end{figure*}

\noindent \textit{\textbf{Finetuning Strategies.}} 
We compare PRE-Net for full-disc precipitation retrieval in Table~\ref{tab:Ab2}. A UNet trained from scratch on full-disc IR data is employed as the baseline (Row 1).
Compared to other models that are trained with MKE, it can be seen as an indicator of whether multimodal knowledge is implied in the fine-tuned model.
In Row 3, we show that the PRE-Net model excelling in swath retrievals fails to generalize to full-disc retrievals, despiting both being IR data. It underscores the necessity of the Full-Disc Adaptation stage for full-disc retrieval.
Row 4-6 achieve better retrieval results than Row 3 further proving this claim. It confirms that these finetuning strategies effectively balance prior multimodal knowledge with full-disc IR knowledge, leading to superior precipitation retrieval.

To further validate the effectiveness of Self-MaskTune and its role in mitigating knowledge forgetting, we compare our Self-MaskTune method with Non-MaskTune and Rand-MaskTune.
Non-MaskTune fine-tunes all parameters without any masking, relying solely on LoRA. Rand-MaskTune uses LoRA for fine-tuning while using randomly generated masks without loss guidance. Self-MaskTune employs a dynamic masking strategy where the mask is guided by prediction errors.
Integrating KD with Non-MaskTune (Row 2) yields improvements over the baseline, which validates the effectiveness of the multimodal knowledge expansion pipeline.
PRE-Net with Non-MaskTune (Row 4) performs worse than KD with Non-MaskTune (Row 2) and even underperforms the baseline (Row 1), indicating that LoRA alone is insufficient for effective adaptation. Without guided updates, LoRA may preserve conflicting multimodal features that misalign with the full-disc IR domain, degrading the model’s ability to learn new patterns. This underscores the need for targeted adaptation, which is provided by Self-MaskTune.
Comparing Row 6, to Row 1, our method (PRE-Net with Self-MaskTune) exhibits superior retrieval performance (2.76\% improvement in RMSE, 2.65\% in CC, 24.27\% in POD, and 11.21\% in CSI).

Note that a slight increase in the FAR metric is acceptable in precipitation retrieval. On the one hand, addressing underreporting is more critical than false reporting, as it could prevent timely responses to potential disasters.
On the other hand, achieving a higher POD often comes at the cost of a higher FAR~\cite{peca}.

\subsection{Comparison with precipitation retrieval products}
\noindent \textit{\textbf{Quantitative Comparison.}}
Table \ref{tab:comparison_methods} quantifies PRE-Net’s superiority over some precipitation products. 
Compared to NWP-based IR methods, PRE-Net significantly outperforms PERSIANN-CCS and PDIR. In classification metrics, PRE-Net achieves notable improvements: POD (+27.46\%), FAR (-8.81\%), and CSI (+17.40\%) over PERSIANN-CCS, and POD (+10.10\%), FAR (-15.12\%), and CSI (+11.31\%) over PDIR. For regression metrics, PRE-Net achieves a lower RMSE (0.8714), with reductions of 0.7344 and 3.3747 compared to PERSIANN-CCS and PDIR, respectively, and a higher CC (0.5028).
These results demonstrate PRE-Net’s ability to 
minimize false alarms while resolving fine-scale precipitation intensity variations.

Notably, PRE-Net demonstrates performance comparable to IMERG, despite IMERG’s reliance on multimodal data and superior computational resources. PRE-Net achieves IR-based retrieval with significantly lower computational costs, faster temporal resolution (5 minutes), and higher spatial resolution (4 km). In classification metrics, PRE-Net reduces FAR (-4.28\%) and increases CSI (+4.30\%) while maintaining a high POD (+4.88\%). In regression metrics, it achieves a lower RMSE (-0.2360) and higher CC (+5.44\%). These results position PRE-Net as a highly viable and cost-effective alternative to IMERG, offering competitive performance with reduced resource demands.
While IMERG exhibits higher CSI-4 (0.4176) and CSI-8 (0.4891) scores, this advantage stems from spatial smoothing during interpolation to 4 km resolution. This inherent spatial smoothing weakens its CSI.
PRE-Net, operating natively at 4 km, achieves competitive scores in CSI-4 (0.4405) and CSI-8 (0.4815) without resolution degradation, confirming its ability to preserve structural fidelity across different spatial resolutions.

\noindent \textit{\textbf{Qualitative Comparison.}}
We visualize and compare the retrieval results across diverse precipitation events in Fig.\ref{fig:fig5}. PRE-Net consistently reconstructs precipitation structures, accurately capturing both spatial distribution and intensity.
(i) Frontal rain. In this strip-shaped precipitation in the central region, PERSIANN-CCS, PDIR, and UNet fail to accurately capture the intensity of this precipitation band. IMERG presents a more refined precipitation intensity but fails to form the shape of frontal rain. PRE-Net closely matches the GT in both distribution and intensity.
(ii) Convective rain. In this heavy precipitation scattered over a large area, PERSIANN-CCS only shows sparse precipitation, PDIR shows weak precipitation intensity, and UNet fails to detect most precipitation. IMERG performs better in distribution while its intensity is overestimated. PRE-Net accurately locates and quantifies heavy rainfall.
(iii) Orographic rain. Heavy concentrations in the Himalayas area. PERSIANN-CCS, PDIR, and UNet completely fail to capture the precipitation. IMERG slightly underestimates the intensity. PRE-Net successfully identifies this orographic rain, aligning closely with the GT.

\begin{table}[t]
    \centering
    \small 
    \setlength{\tabcolsep}{0.6mm} 
    \renewcommand{\arraystretch}{1.5} 
    \begin{tabular}{@{\hspace{0mm}}c|ccccc@{\hspace{0mm}}}
        \toprule
         \textbf{Method} & \textbf{RMSE $\downarrow$} & \textbf{CC $\uparrow$} & \textbf{POD $\uparrow$} & \textbf{FAR $\downarrow$} & \textbf{CSI $\uparrow$} \\
        \midrule
        PERSIANN-CCS & 11.1723 & 0.1204 & 0.3083 & 0.0113 & 0.3073 \\
        PDIR & 11.1150 & 0.1280 & \textbf{0.4113} & 0.0090 & \textbf{0.4098} \\
        IMERG & 11.0625 & 0.1738 & 0.3650 & 0.0118 & 0.3634 \\
        \midrule
        PRE-Net 
        & \textbf{11.0303} & \textbf{0.1989} & 0.3906 & \textbf{0.0068} & 0.3896 \\
        \bottomrule
    \end{tabular}
    \caption{Regional Generalization Results over Australia (Wet Season 2022–2023).}
    \vspace{-2em}
    \label{tab:australia}
\end{table}

\noindent \textit{\textbf{Regional Generalization.}}
While the selected region already covers diverse precipitation regimes from tropical to temperate zones, we further evaluate the generalization capability of PRE-Net. Specifically, we conducted experiments over Australia (10–40°S, 113–153°E) during 2022-12-01 to 2023-02-28 wet season. All evaluations are conducted using station-level ground truth data from the Integrated Surface Dataset (ISD) maintained by the National Oceanic and Atmospheric Administration (NOAA). Compared with IMERG, PRE-Net outperforms IMERG across all key metrics, as shown in \ref{tab:australia}. These results demonstrate PRE-Net's potential for global deployment.

\begin{table}[t]
    \centering
    \small 
    \setlength{\tabcolsep}{0.6mm} 
    \renewcommand{\arraystretch}{1.5} 
    \begin{tabular}{@{\hspace{0mm}}c|ccccc@{\hspace{0mm}}}
        \toprule
         \textbf{Noise injection} & \textbf{RMSE $\downarrow$} & \textbf{CC $\uparrow$} & \textbf{POD $\uparrow$} & \textbf{FAR $\downarrow$} & \textbf{CSI $\uparrow$} \\
        \midrule
        - & 0.8714 & 0.5028 & 0.5597 & 0.4349 & 0.3912\\
        Additive Noise & 0.8751 & 0.4999 & 0.5198 & 0.4256 & 0.3753 \\
        Multiplicative Noise & 0.8731 & 0.5023 & 0.5530 & 0.4315 & 0.3895 \\
        \bottomrule
    \end{tabular}
    \caption{Robustness of PRE-Net under Gaussian Noise Perturbations.}
    \vspace{-2em}
    \label{tab:noise}
\end{table}

\noindent \textit{\textbf{Noise Sensitivity.}}
In practical retrieval systems, satellite data are frequently affected by various types of noise, such as atmospheric interference, sensor calibration drift. Such perturbations may degrade retrieval performance and compromise reliability in operational settings. Therefore, we evaluate the noise sensitivity of PRE-Net by injecting synthetic noise(additive Gaussian noise and multiplicative Gaussian noise) into the input data. As shown in \ref{tab:noise}, PRE-Net demonstrates remarkably stable performance despite the presence of noise, indicating PRE-Net's resilience to noise.

\noindent \textit{\textbf{Operational Efficiency.}}
Combining 4 km resolution, 5-minute updates, and IR-only inputs, PRE-Net offers a computationally efficient alternative to SPPs' resource-intensive multimodal pipelines. Its balance of accuracy and operational feasibility positions it as a promising tool for real-time, high-resolution precipitation monitoring.


\section{Conclusion}
We propose the Precipitation Retrieval Expansion (PRE) task in this paper, which aims to achieve high-quality IR-based full-disc precipitation retrievals beyond the scanning swath. To address the PRE task, we develop a two-stage Multimodal Knowledge Expansion pipeline and the PRE-Net model, consisting of Swath-Distilling for knowledge transferring from multimodal to IR and Full-Disc Adaptation stages for generalization from scanning swath to full-disc precipitation retrieval. In the PRE-Net model, we introduce the Coordinated Masking and Wavelet Enhancement module to effectively encode richer knowledge. For the Full-Disc Adaptation stage, we develop Self-MaskTune to smoothly focus on disagreements between distilled multimodal knowledge and new IR-only knowledge, mitigating the impact of knowledge disparity, facilitating domain transfer for scalable predictions. Extensive experiments and analyses demonstrate that PRE-Net is an effective solution for the PRE task.

\vspace{0.4em}

\section*{Acknowledgements}
This work is partially supported by Zhejiang Provincial Natural Science Foundation of China under Grant No. LRG25F020002, LR21F020002, the project of China Meteorological Service Association under NO. CMSA2023MD001, Natural Science Foundation of China (No. 62302453), Zhejiang Provincial Natural Science Foundation of China (No. LMS25F020003) and China Meteorological Administration Key Innovation Team (CMA2023ZD01).

\clearpage
\bibliographystyle{ACM-Reference-Format}
\bibliography{sample-base}
\clearpage

\appendix
\onecolumn
\clearpage

\section{Details of the Teacher Network}\label{teacher_details}
Here we present detailed information on the teacher network used in the Swath-Distilling stage. The teacher is trained with multimodal data for scanning swath precipitation retrievals, built upon the foundational U-Net framework. The network processes the integrated multimodal data (i.e., IR, PMW, and PR reflectivity) as inputs. Down-sampling in the encoder consists of a double convolutional block followed by 2×2 max pooling. The decoder, responsible for up-sampling, mirrors this structure. The initial number of feature layers starts at 64, doubling with each down-sampling step and halving with each up-sampling. Additionally, we incorporate geographical data (elevation, latitude, and longitude) late in the network to address the terrain's influence on precipitation. 

We implement the teacher model with training and inference conducted on an NVIDIA RTX A6000 GPU. The model is optimized using RMSprop with an initial learning rate of 1e-6, which is halved every 30 epochs. The training is performed over 200 epochs with a batch size of 32. CLDAS-V2.0 is used as the ground truth due to its superior quality in China compared to similar international products. The teacher model is validated every 20 epochs, and the best-performing model on the validation set is selected for evaluation on the test set.

The comparison of performance metrics in Tables 8 and 9 demonstrates the efficacy of integrating multiple data sources (IR, PMW, and PR) for hourly precipitation retrieval. In Table 8, the combination of all three data types consistently yields the best performance across all metrics, with the lowest RMSE (0.7290 mm/h) and the highest CSI (0.4524), indicating a more accurate and reliable precipitation retrieval. Notably, the inclusion of PR data significantly improves performance, especially in reducing the FAR and enhancing the CC, suggesting that PR data contributes substantially to the accuracy of precipitation detection. Meanwhile, the results highlight the limitations of using IR data alone for precipitation retrieval, as it yields the highest RMSE (0.9781 mm/h) and the lowest CC (0.3408), POD (0.2159), and CSI (0.1831) among all configurations. This suggests that IR data lacks the ability to accurately capture precipitation patterns on its own, and its combination with other data types (PMW and PR) is essential for improving retrieval accuracy and reducing errors.

Table 9 compares the performance of the teacher model (using all three data types) with the GPM 2B-CMB product. The teacher model outperforms GPM 2B-CMB across all metrics, with a notable improvement  (41.27\% improvement in RMSE, 25.30\% in CC, 8.96\% in POD, 8.84\% in FAR and 9.32\% in CSI), demonstrating its superior ability to detect precipitation events accurately. This suggests that the teacher model, which leverages multi-modal data, is more effective in capturing the complexities of precipitation patterns than the GPM 2B-CMB product.

\begin{table*}[h]
\centering
\setlength{\tabcolsep}{3mm}
\renewcommand{\arraystretch}{1.4}{%
\begin{tabular}{@{}cccccccc@{}}
\toprule
IR           & PMW          & PR          & RMSE(mm/h) ↓ & CC ↑    & POD ↑    & FAR ↓             & CSI ↑    \\ \midrule
$\checkmark$ &              &              & 0.9781   & 0.3408 & 0.2159 & 0.4536          & 0.1831 \\
             & $\checkmark$ &              & 0.7545	&0.6824	&0.5279	&0.2678	&0.4425 \\
             &              & $\checkmark$ & 0.7383	&0.7040	&0.5409	&0.2605	&0.4544 \\ \midrule
$\checkmark$ & $\checkmark$ &              &0.7684	&0.6705	&0.5295	&0.2917	&0.4347 \\
$\checkmark$ &              & $\checkmark$ & 0.7351	&0.7048	&0.4853	&\textbf{0.2273}	&0.4274 \\
             & $\checkmark$ & $\checkmark$ &\textbf{ 0.7290}	&0.7076	&0.5245	& 0.2330	&0.4524 \\ \midrule
$\checkmark$ & $\checkmark$ & $\checkmark$ & 0.7436 & \textbf{0.7213} & \textbf{0.5511} & 0.2358 & \textbf{0.4710} \\ \bottomrule
\end{tabular}%
}
\caption{Ablations of hourly precipitation retrieval performance from different modalities.}
\end{table*}

\begin{table*}[h]
\centering
\setlength{\tabcolsep}{3mm}
\renewcommand{\arraystretch}{1.4}{%
\begin{tabular}{lccccc}
\hline
           & RMSE(mm/h) ↓     & CC ↑              & POD ↑             & FAR ↓    & CSI ↑             \\ \toprule
Teacher   & \textbf{0.7436}& \textbf{0.7213}& \textbf{0.5511}& \textbf{0.2358}& \textbf{0.4710}\\
GPM 2B-CMB & 1.1563 & 0.4683 & 0.4615 & 0.3242 & 0.3778          \\ \bottomrule
\end{tabular}%
} 
\caption{Performance compared with GPM 2B-CMB on Swath-MPR.}
\end{table*}


\end{document}